\documentclass[11pt]{article}
\usepackage[final]{acl}
\usepackage{times,latexsym}
\usepackage[T1]{fontenc}
\usepackage[utf8]{inputenc}
\usepackage{microtype,amsmath,amssymb,booktabs,graphicx,colortbl}
\definecolor{resultshade}{RGB}{236,241,245}
\newcommand{\tablefont}{\small}
\title{Scalar Tuned Lens Summaries Offer Limited Incremental Value \\for Mid-Generation Abstention}
\author{Khurram Khalil \and Khaza Anuarul Hoque \\
University of Missouri-Columbia, MO, USA \\
\texttt{\{khurram.khalil,hoquek\}@missouri.edu}}
\begin{document}
\maketitle
\begin{abstract}
Mid-generation abstention aims to stop an unpromising response before spending its remaining computation budget. Intermediate transformer states may help predict eventual correctness, but whether scalar summaries of their vocabulary predictions add useful evidence is less clear. We compare Tuned Lens entropy and divergence features with final-state and raw intermediate-state probes under matched, causal evaluation. Across eight instruction-tuned models on MATH-500, raw intermediate states improve ranking, whereas scalar-only probes usually trail the final-state baseline. Scalar hybrids have positive exceptions, but their advantage over position-matched permutation controls is small. Cross-fitted residual comparisons further show that gains over a final-state probe need not require aligned scalar features. Actual stopping runs on two model sizes and two benchmarks reveal a separate accuracy--computation trade-off, with token savings overstating wall-time savings for scalar policies. The evidence supports a restricted empirical conclusion about these scalar features, not a general negative claim about intermediate representations.
\end{abstract}

\section{Introduction}
Additional reasoning tokens are useful only when they improve a response enough to justify their cost. A model can continue an incorrect argument for hundreds of tokens without recovering, so a predictor of eventual correctness may support early abstention. Unlike post-generation rejection, this decision must rely entirely on the prefix already generated and must pay its own inference cost. A useful predictor therefore needs more than a favorable retrospective ranking score.

\citet{davidov2026quit} study dynamic abstention through a value estimate of eventual correctness. Their final-layer hidden-state probe provides a natural reference for asking whether other internal representations improve the decision. The final state summarizes the current prefix, but it does not explicitly expose the sequence of intermediate vocabulary predictions. The Tuned Lens maps intermediate states through learned affine translators into vocabulary distributions \citep{belrose2023tunedlens}. Entropy and divergence may reveal instability.

This motivation leaves two different empirical questions. First, do intermediate states carry predictive signal that a final-state probe fails to extract? Second, do the particular scalar summaries add useful signal beyond a matched final-state baseline? A negative answer to the second question cannot answer the first. Conversely, a larger hybrid probe may improve even when the extra input dimensions contain no correctly aligned trajectory. Raw-state, position, and permutation controls test these gains.

\begin{figure*}[t]
\centering
\includegraphics[width=\textwidth]{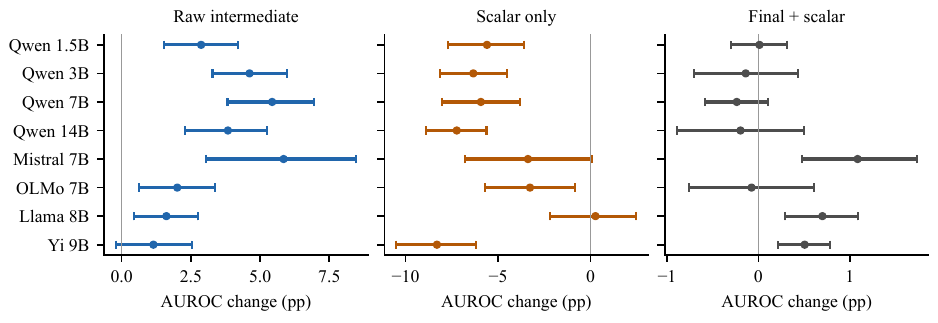}
\caption{\textbf{Raw states and scalar summaries give different ranking results.} Each point is a change in problem-balanced token ROC-AUC relative to the final-state probe on 500 MATH-500 problems for one model, averaged across three fitting seeds and five outer folds. Bars are paired 95\% problem-bootstrap intervals conditional on the fitted probes. Raw intermediate states, scalar-only features, and final-state plus scalar features appear in separate panels. Zero denotes no change. Individual intervals are not simultaneous multiplicity-adjusted intervals.}
\label{fig:ranking}
\end{figure*}

\section{Task and Representations}
\label{sec:method}
Let $y_{1:t}$ be the generated prefix for problem $x$, and let $c\in\{0,1\}$ indicate whether a completed response is correct under the specified grader. A probe estimates $c$ from the current prefix representation. It is a learned predictor, not the true continuation value. Prediction errors and extraction costs limit policy guarantees.

For an $L$-block transformer, let $h_{d,t}$ denote the state after $d$ blocks. In our extraction convention, $h_{0,t}$ is the embedding state and $h_{L,t}$ includes the final normalization. Translator $T_d$ consumes the state after $d$ blocks, for $d=0,\ldots,L-1$. With final normalization $N$ and unembedding matrix $E$, its distribution is $p_{d,t}=\operatorname{softmax}(E N(T_d h_{d,t}))$.
The final model distribution is $p_{L,t}=\operatorname{softmax}(E h_{L,t})$. The scalar vector concatenates $H(p_{d,t})$ and $D_{\mathrm{KL}}(p_{d,t}\Vert p_{L,t})$ over measured depths. Thus the divergence reference is the final distribution, not the adjacent layer. Computing this vector requires all measured vocabulary projections and the final model state. It does not permit an early exit before the final block.

The final-state baseline receives $h_{L,t}$. The raw intermediate baseline receives $h_{m,t}$ at a fixed depth near 60\% of the blocks. Scalar-only probes receive the $2L$ scalar vector; hybrid probes concatenate that vector with $h_{L,t}$. A final-plus-intermediate comparison instead concatenates $h_{L,t}$ and $h_{m,t}$. Every primary variant also receives the same causal position features. No reference-answer token probability enters these predictors.

The position vector contains $t/B$, $\log(1+t)/\log(1+B)$, $\sin(\pi t/B)$, and $\cos(\pi t/B)$, with a fixed budget $B=2048$. These quantities are available during generation. The completed response length is never an input. Removing position from both final-state and hybrid probes provides a separate sensitivity check; a position-only probe tests how much progress alone predicts correctness.

For residual comparisons, a correction is added to the frozen baseline logit. The correction is trained against baseline predictions obtained by three-way cross-fitting within the outer training problems. Its inputs are either scalar features with position, position alone, or a permuted scalar control. This avoids fitting the correction against baseline training predictions from a model that saw those same labels. Finite tuning data still introduce model-selection uncertainty.

\section{Evaluation Protocol}
\label{sec:protocol}
The primary ranking results, incremental controls, and operational outcomes are reported in the main text. Supplementary Appendices~\ref{app:models}--\ref{app:cost} provide checkpoint and layer details, sensitivity definitions, historical corrections, and cost conventions. Appendix~\ref{app:details} gives training settings and results.

The eight models are Qwen2.5-Instruct at 1.5B, 3B, 7B, and 14B \citep{yang2024qwen25}; Mistral-7B-Instruct-v0.3 \citep{jiang2023mistral}; OLMo-2-1124-7B-Instruct \citep{teamolmo2024olmo2}; Llama-3.1-8B-Instruct \citep{grattafiori2024llama3}; and Yi-1.5-9B-Chat \citep{young2024yi}. Their primary benchmark is the 500-problem MATH subset associated with \citet{lightman2024verify}, drawn from MATH \citep{hendrycks2021math}. The online study additionally uses MMLU-Pro \citep{wang2024mmlupro}. A reasoning-trained checkpoint is evaluated separately as an exploratory extension.

Historical response caches do not retain exact generated token identifiers for all instruction models. Offline features are therefore extracted from reconstructed prompts and retokenized response text. The resulting ranking and replay experiments describe that reconstruction and cannot establish exact agreement with the historical generator. Our online study instead saves the actual prompt and generated token identifiers and checks each stopping decision on those exact prefixes.

Mathematical grading extracts a final candidate without consulting the reference and applies symbolic comparison with Math-Verify \citep{kydlicek2025mathverify}. A partial-expression check rejects a matching numeric substring of a larger incorrect expression. The corrected implementation preserves all nine current cached label vectors. Historical regrading yields 412 incorrect-to-correct and one correct-to-incorrect transition, totaling 413 changes and a net increase of 411. Appendix~\ref{app:corrections} distinguishes these counts from comparisons involving stored labels. Automated labels remain uncertain.

\subsection{Splits and model selection}
Five outer folds split problems, never individual tokens. Within each outer fold, 20\% of the non-test problems are reserved for tuning, then 25\% of the remainder for threshold calibration. For 500 problems this gives 240 training, 80 tuning, 80 calibration, and 100 test problems. Splits are stratified by label and fixed across fitting seeds and variants within a model. The labels differ across models, so stratified folds are not asserted to share identical test membership across models.

\subsection{Ranking and uncertainty}
The primary metric is problem-balanced token ROC-AUC. Each sampled position receives weight inversely proportional to its problem's number of sampled positions. Problems without eligible positions contribute to policy evaluation but not token ranking. Ordinary token ROC-AUC, Brier score, and prefix-ranking risk--coverage area are separate outputs. In particular, ROC-AUC is not area under the selective-accuracy curve.

Paired bootstrap intervals resample held-out problems and average differences across the three fitted seeds within each resample. We use 2000 resamples. These intervals condition on the fitted models and the available benchmark; they do not capture full training-population uncertainty or make repeatedly inspected MATH-500 problems fresh confirmation data. Related Qwen scales are not independent family replications.

For incremental comparisons we report Holm correction \citep{holm1979simple} across four contrasts on eight models, giving 32 tests. The contrasts are hybrid minus final, residual minus final, hybrid minus permuted, and final-plus-intermediate minus final. This family is specified for the reporting analysis, not claimed as preregistered. A 0.01 AUROC practical margin, with 0.005 and 0.02 sensitivities, distinguishes small effects from mere nonsignificance. Individual confidence bounds are not simultaneous equivalence guarantees.

Features are standardized on training problems. Probe selection compares linear heads and hidden widths 128 and 256 with weight decay $10^{-4}$ or $10^{-2}$. Cross-fitted residual sensitivity separately selects output penalties from 0, 0.01, 0.1, and 1. AdamW uses learning rate 0.001, batch size 8192, and at most 40 epochs with five-epoch early-stopping patience. Selection minimizes tuning binary cross-entropy. Problem weighting prevents longer responses from dominating the token loss. For the residual-penalty sensitivity, the previously selected architecture is held fixed. Equal tuning grids therefore provide comparable selection budgets while allowing different input dimensions and parameter counts. Further settings appear in Appendix~\ref{app:details}.

\section{Ranking and Incremental Value}
\subsection{Raw intermediate states retain useful signal}
Table~\ref{tab:ranking} and Figure~\ref{fig:ranking} separate raw intermediate states from the scalar representation. The raw-state improvement ranges from 1.15 to 5.85 AUROC points across the eight primary models. Seven individual 95\% intervals exclude zero; the Yi interval crosses zero. The depth was selected during earlier exploratory work and held fixed here. These results demonstrate predictive signal at the tested depth. Depth optimality remains untested.

Scalar-only differences range from $-8.30$ points on Yi to $+0.27$ on Llama. The Llama interval spans both signs, as does the Mistral interval. Consequently, a claim that scalar-only probes significantly underperform on every model is unsupported by this corrected comparison. The consistent practical issue is that these summaries do not reproduce the raw-state gains. This distinction limits the conclusion to a representation and probe family rather than to intermediate processing generally.

\subsection{Positive hybrid exceptions remain}
Table~\ref{tab:incremental} reports the incremental comparisons directly. Hybrid-minus-final gains survive Holm correction on Mistral, Llama, and Yi, with point estimates of 1.08, 0.70, and 0.51 AUROC points. These positive exceptions contradict an unconditional no-gain statement and are retained in the interpretation. The other five models have small changes whose individual intervals span zero.

The aligned-versus-permuted hybrid differences are smaller. No such comparison passes the specified 32-test family. Permutation is performed within causal position and separately within training, tuning, calibration, and test partitions. The final-state and position inputs remain aligned. This comparison tests incremental dependence on the aligned scalar block under a particular conditional shuffle; it is not a proof that any added dimensions are harmless or that position carries no signal.

Five additional permutation draws assess sensitivity to the shuffle. Random added features also improve on the final-state baseline for Mistral, Llama, and Yi by 1.15, 0.51, and 0.51 AUROC points. These comparisons qualify attribution of hybrid gains to aligned scalar trajectories. Matching the intermediate probe to the final-state-selected architecture leaves six of eight individual intervals above zero. Appendix~\ref{app:sensitivity} lists all intervals.

\subsection{Residual gains depend on the comparator}
Cross-fitted scalar residual corrections improve over the final-state baseline for several models. Four pass the 32-test family, so residual fitting does not support a blanket absence-of-gain claim either. However, a residual correction also changes the functional class and recalibrates the baseline. A position-only correction tests this explanation independently of scalar features.

After separate output-penalty selection, scalar residual corrections have lower point estimates than position-only corrections for all eight models. Their differences range from approximately $-1.81$ points on Mistral to $-0.005$ on Qwen 3B. The largest individual upper 95\% bound is 0.21 points, below the stated 0.5-, 1-, and 2-point practical margins. This is evidence against a large scalar advantage over this comparator under the fixed protocol, not simultaneous equivalence over all models or all possible residual architectures. Appendix~\ref{app:sensitivity} gives the intervals and permutation ranges.

Math-trained lenses retain positive hybrid exceptions on Mistral and Llama among six tested models. Appendix~\ref{app:sensitivity} gives the intervals and permutation controls. These comparisons assess prediction with math-trained lenses, without establishing superiority over web-trained lenses.

\section{Actual Stopping and Measured Cost}
\label{sec:online}
\begin{figure*}[t]
\centering\includegraphics[width=\textwidth]{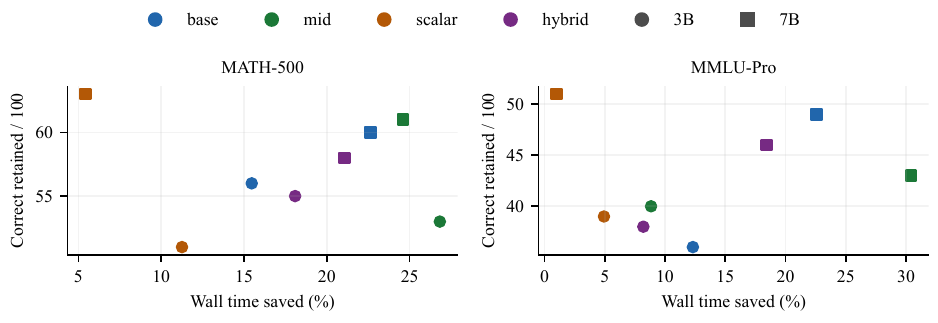}
\caption{\textbf{Measured time savings and retained correct answers.} Each point is one online policy on 100 preselected held-out problems, with Qwen2.5-3B shown by circles and 7B by squares. Colors identify final-state (base), raw intermediate-state (mid), scalar-only, and hybrid policies. Thresholds target 75\% calibration coverage with patience one; achieved test coverage differs. Higher and farther right is preferable on these two axes, but points do not form an optimized frontier. Table~\ref{tab:online} supplies uncertainty and coverage.}
\label{fig:policy}
\end{figure*}

\subsection{Causal stopping protocol}
Policies check after every 16 generated tokens, beginning at token 16. A response is withheld once its score falls below a calibration-selected threshold for the required number of consecutive checks. Patience is evaluated at the last required check, never retroactively. The primary online comparison uses patience one and target calibration coverage 75\%; replay additionally evaluates patience three and targets 90\%, 50\%, and 25\%. Thresholds are frozen before evaluating test problems. Achieved test coverage need not equal the calibration target.

The online study contains 100 preselected outer-fold-zero test problems for each of four settings, Qwen 3B and 7B on MATH-500 and MMLU-Pro. Each setting is split into two 50-problem shards on separate A100 80 GB GPUs. Full generation, final-state, intermediate-state, scalar-only, and hybrid policies run on every problem, with deterministic rotation of policy order. Generation is greedy with repetition penalty 1.1 and a maximum of 2048 new tokens. Batch size is one. Warm-up is excluded and CUDA execution is synchronized around timing.

Table~\ref{tab:online} reports achieved coverage, accuracy among retained responses, correct answers retained per original problem, generated-token savings, and measured wall-time savings. The last two quantities are ratios of aggregate counts or times, not averages of per-problem ratios. Coverage denotes the fraction of original problems receiving an answer. Retained accuracy is the fraction of those answers graded correct. Correct answers per original problem equals their product, allowing comparisons that account for both abstention and correctness. Withholding any response lowers coverage. Withholding a correct response also lowers answer retention.

On MATH-500 Qwen 3B, the final-state policy retains 56 correct answers per 100 problems and saves 15.5\% wall time. The intermediate policy retains 53 and saves 26.8\%. Calling the latter uniformly superior would ignore three additional correct answers withheld. On Qwen 7B, the intermediate policy retains 61 correct answers versus 60 for the final-state policy and saves 24.6\% versus 22.6\% wall time. These are point comparisons at calibration-selected thresholds, not statistically established Pareto dominance or a comparison at identical achieved coverage.

Scalar-only policies yield particularly different token and time savings. On MMLU-Pro Qwen 7B, token savings are 6.2\%, while measured time savings are 1.0\% with an interval from $-2.5$ to $4.6$\%. That interval does not establish a positive speedup. The same scalar policy retains 51 correct answers per 100 problems, compared with 49 for the final-state policy, while accepting more responses. Thus the scalar outcome is not uniformly worse on every axis, even though its timing benefit is uncertain.

Vocabulary projections contribute materially to scalar cost. Including every measured layer adds approximately 22.5\% and 13.6\% dense FLOPs for Qwen 3B and 7B at context length 512 with checks every 16 tokens. Timing includes prefill, decoding, probes, and lens extraction. All 400 online cases passed exact-prefix and replay checks. The intermediate policy can stop before later blocks execute. Scalar timing includes vocabulary projections and auxiliary probe scores. Matched-coverage and matched-error comparisons remain open.

\section{Related Work and Scope}
The value-based formulation of \citet{davidov2026quit} motivates predicting correctness during generation. Selective classification studies accuracy among retained examples \citep{geifman2017selective}. Our evaluation measures answer retention and computation across achieved coverage levels.

The Tuned Lens decodes latent predictions from transformer states \citep{belrose2023tunedlens}. Its entropy and divergence summaries may have different predictive value from the underlying states. Our raw-state and permutation controls examine this distinction for abstention. \citet{lavi2026unanswerability} instead study whether an extractive question-answering context supports an answer. Our labels concern correctness of a particular completed response.

Intermediate-state abstention withholds a response. Layer-adaptive methods such as DeeBERT \citep{xin2020deebert} and confident adaptive language modeling \citep{schuster2022calm} pursue related computation savings with different output objectives. Token counts and measured wall time remain separate outcomes in our evaluation.

\section{Conclusion}
Across eight models, raw intermediate states improve correctness prediction by point estimate. Scalar hybrids and residual corrections have positive exceptions, while matched controls qualify attribution of these gains to aligned scalar trajectories. Online stopping further exposes trade-offs among time saved, coverage, and correct answers retained. These findings support limited incremental value for the tested scalar summaries.

\section*{Acknowledgement}
This work used computational resources from NCSA Delta GPU and Purdue Anvil AI at UIUC/NCSA and Purdue University, respectively, through allocation CIS260364 from the Advanced Cyberinfrastructure Coordination Ecosystem: Services \& Support (ACCESS) program, supported by U.S. National Science Foundation (NSF) Grants 2138259, 2138286, 2138307, 2137603, and 2138296. Additional GPU resources (8 NVIDIA A100 80 GB GPUs on the Brev platform) were provided by the NVIDIA Academic Grant Program.

The authors would like to acknowledge the use of Grammarly (an AI-assisted editing tool), for language editing, proofreading, and typographical corrections during the preparation of this manuscript. All technical content, analyses, and conclusions were manually drafted and verified by the authors.

\clearpage

\section*{Limitations}
\label{sec:limitations}
Mathematical labels are automatically assessed. The 125 current-grader disagreements with stored labels and 180 sampled agreements have not undergone complete human adjudication. MATH-500 has been repeatedly inspected during development. The fixed intermediate depth comes from prior exploratory work. Conditional bootstrap intervals omit full training-population uncertainty and do not certify simultaneous equivalence across models.

Offline reconstruction may differ from the original prompt and token sequence. Exact online verification applies to the four new settings, not retroactively to every cached trace. Online generation uses two Qwen sizes, one hardware configuration, batch size one, and one paired repetition per problem. Policy order rotates, separate GPUs share the host, and scalar overhead includes auxiliary scoring. The study does not establish optimized serving throughput, a universal abstention reward, or superiority at equal achieved coverage.

The feature set is narrow, and finite tuning budgets limit negative conclusions. The residual-penalty sweep fixes previously selected architecture choices. Whole-problem permutations condition on observed schedules. We do not establish causal separation of domain shift from class balance or an information-theoretic loss from scalar compression. Historical experiments and their total GPU hours were not consistently recorded; current completion does not repair that historical accounting.

\section*{Ethical Considerations}
Abstention may withhold useful correct responses while leaving confidently incorrect ones. Probe scores should not be presented as guarantees of reliability, especially under distribution shift or in consequential applications. The study uses existing benchmark responses rather than new human-subject data. Checkpoint and dataset licenses remain applicable to redistribution. Automatically graded answers carry residual labeling uncertainty.

\bibliography{references}
\appendix

\section{Detailed Results and Probe Selection}
\label{app:details}

Features are standardized using the training partition alone, with standard deviation clipped below at 0.01. Candidate probes are linear logistic heads or one-hidden-layer ReLU networks with width 128 or 256. AdamW uses learning rate 0.001, batch size 8192, and weight decay selected from $10^{-4}$ and $10^{-2}$. Training lasts at most 40 epochs with five-epoch early-stopping patience. Probe selection minimizes tuning binary cross-entropy, not test AUROC. Three fitting seeds are retained. Token losses are weighted so that long responses do not dominate solely through more sampled positions.

Residual-output penalties are evaluated separately at 0, 0.01, 0.1, and 1. That sensitivity holds the previously selected architecture fixed and selects the penalty on tuning loss. It is therefore not an exhaustive joint architecture--penalty search. An additional raw-intermediate comparison matches the final-state architecture choice. Equal search grids should not be mistaken for equal parameter counts when input dimensions differ.

\begin{table*}[t]
\centering\tablefont\setlength{\tabcolsep}{4pt}
\begin{tabular}{lrrr}
\toprule
\textbf{Model} & \textbf{Final AUROC} & \textbf{Raw intermediate minus final} & \textbf{Scalar minus final} \\
\midrule
\rowcolor{resultshade}
Qwen 1.5B & 0.690 & $+2.87$ [+1.53, +4.19] & $-5.60$ [-7.71, -3.57] \\
Qwen 3B & 0.729 & $+4.62$ [+3.28, +5.97] & $-6.33$ [-8.12, -4.52] \\
\rowcolor{resultshade}
Qwen 7B & 0.725 & $+5.43$ [+3.82, +6.96] & $-5.93$ [-8.02, -3.83] \\
Qwen 14B & 0.763 & $+3.84$ [+2.29, +5.24] & $-7.23$ [-8.88, -5.62] \\
\rowcolor{resultshade}
Mistral 7B & 0.640 & $+5.85$ [+3.05, +8.46] & $-3.38$ [-6.80, +0.07] \\
OLMo 7B & 0.702 & $+2.01$ [+0.63, +3.36] & $-3.27$ [-5.69, -0.84] \\
\rowcolor{resultshade}
Llama 8B & 0.689 & $+1.62$ [+0.43, +2.76] & $+0.27$ [-2.18, +2.48] \\
Yi 9B & 0.756 & $+1.15$ [-0.20, +2.53] & $-8.30$ [-10.49, -6.21] \\
\bottomrule
\end{tabular}

\caption{\textbf{Primary ranking comparisons on MATH-500.} Final AUROC is the mean problem-balanced token ROC-AUC over three fitting seeds, with each problem tested once per seed across five folds. Differences are AUROC percentage points with paired 95\% problem-bootstrap intervals. All probes receive the same causal position inputs and tuning grid. The raw intermediate depth is fixed near 60\% of model depth. Shading alternates rows for readability and does not denote significance.}
\label{tab:ranking}
\end{table*}

\begin{table*}[t]
\centering\tablefont\setlength{\tabcolsep}{3.5pt}
\begin{tabular}{lrrr}
\toprule
\textbf{Model} & \textbf{Hybrid minus final} & \textbf{Hybrid minus permuted} & \textbf{Final + mid minus final} \\
\midrule
\rowcolor{resultshade}
Qwen 1.5B & $+0.01$ [-0.30, +0.31] & $+0.15$ [-0.12, +0.39] & $+2.91$ [+2.05, +3.78] \\
Qwen 3B & $-0.14$ [-0.70, +0.43] & $-0.04$ [-0.33, +0.25] & $+3.71$ [+2.92, +4.52] \\
\rowcolor{resultshade}
Qwen 7B & $-0.24$ [-0.58, +0.11] & $-0.26$ [-0.56, +0.08] & $+4.45$ [+3.34, +5.50] \\
Qwen 14B & $-0.20$ [-0.89, +0.50] & $-0.26$ [-0.88, +0.32] & $+2.51$ [+1.55, +3.41] \\
\rowcolor{resultshade}
Mistral 7B & $+1.08$ [+0.48, +1.73] & $+0.27$ [-0.00, +0.56] & $+0.38$ [-1.46, +2.10] \\
OLMo 7B & $-0.07$ [-0.76, +0.61] & $+0.11$ [-0.09, +0.30] & $+0.45$ [-0.43, +1.33] \\
\rowcolor{resultshade}
Llama 8B & $+0.70$ [+0.29, +1.08] & $+0.23$ [-0.03, +0.49] & $+1.92$ [+0.79, +3.03] \\
Yi 9B & $+0.51$ [+0.21, +0.79] & $+0.04$ [-0.11, +0.20] & $+1.89$ [+1.14, +2.67] \\
\bottomrule
\end{tabular}

\caption{\textbf{Incremental representation comparisons.} Differences are problem-balanced token ROC-AUC percentage points on MATH-500 with paired 95\% problem-bootstrap intervals, three seeds, and five folds. Hybrid concatenates final states and scalar summaries. Permuted preserves final states and causal position while shuffling scalar blocks within partition and position. Final + mid concatenates final and raw intermediate states. These individual intervals are distinct from the 32-test Holm decisions in the text.}
\label{tab:incremental}
\end{table*}

\begin{table*}[t]
\centering\tablefont\setlength{\tabcolsep}{3pt}
\begin{tabular}{llrrrrr}
\toprule
\textbf{Setting} & \textbf{Policy} & \textbf{Coverage} & \textbf{Accuracy} & \textbf{Correct/all} & \textbf{Tokens saved} & \textbf{Time saved [95\% CI]} \\
\midrule
\rowcolor{resultshade}
MATH 3B & full & 100.0 & 60.0 & 60.0 & 0.0 & 0.0 [0.0, 0.0] \\
MATH 3B & base & 79.0 & 70.9 & 56.0 & 15.2 & 15.5 [8.7, 22.5] \\
\rowcolor{resultshade}
MATH 3B & mid & 70.0 & 75.7 & 53.0 & 26.2 & 26.8 [18.6, 34.9] \\
MATH 3B & scalar & 71.0 & 71.8 & 51.0 & 14.4 & 11.3 [5.7, 16.7] \\
\rowcolor{resultshade}
MATH 3B & hybrid & 73.0 & 75.3 & 55.0 & 21.1 & 18.1 [9.7, 25.8] \\
MATH 7B & full & 100.0 & 71.0 & 71.0 & 0.0 & 0.0 [0.0, 0.0] \\
\rowcolor{resultshade}
MATH 7B & base & 73.0 & 82.2 & 60.0 & 23.2 & 22.6 [14.5, 30.2] \\
MATH 7B & mid & 72.0 & 84.7 & 61.0 & 25.2 & 24.6 [16.4, 32.6] \\
\rowcolor{resultshade}
MATH 7B & scalar & 78.0 & 80.8 & 63.0 & 11.3 & 5.4 [0.7, 9.9] \\
MATH 7B & hybrid & 69.0 & 84.1 & 58.0 & 26.1 & 21.1 [12.6, 28.8] \\
\rowcolor{resultshade}
MMLU-Pro 3B & full & 100.0 & 42.0 & 42.0 & 0.0 & 0.0 [0.0, 0.0] \\
MMLU-Pro 3B & base & 80.0 & 45.0 & 36.0 & 11.9 & 12.3 [7.4, 17.8] \\
\rowcolor{resultshade}
MMLU-Pro 3B & mid & 88.0 & 45.5 & 40.0 & 8.8 & 8.8 [4.1, 14.1] \\
MMLU-Pro 3B & scalar & 76.0 & 51.3 & 39.0 & 8.7 & 4.9 [0.9, 9.1] \\
\rowcolor{resultshade}
MMLU-Pro 3B & hybrid & 81.0 & 46.9 & 38.0 & 12.1 & 8.2 [2.7, 14.3] \\
MMLU-Pro 7B & full & 100.0 & 59.0 & 59.0 & 0.0 & 0.0 [0.0, 0.0] \\
\rowcolor{resultshade}
MMLU-Pro 7B & base & 70.0 & 70.0 & 49.0 & 22.7 & 22.6 [15.4, 30.2] \\
MMLU-Pro 7B & mid & 62.0 & 69.4 & 43.0 & 31.0 & 30.4 [22.3, 39.0] \\
\rowcolor{resultshade}
MMLU-Pro 7B & scalar & 77.0 & 66.2 & 51.0 & 6.2 & 1.0 [-2.5, 4.6] \\
MMLU-Pro 7B & hybrid & 67.0 & 68.7 & 46.0 & 23.4 & 18.4 [11.0, 26.5] \\
\bottomrule
\end{tabular}

\caption{\textbf{Actual online outcomes, not cached-trajectory estimates.} Each setting contains 100 problems, two fixed GPU shards, and one fitted seed and outer fold. All entries except setting and policy are percentages. Accuracy is conditional on retained responses; Correct/all counts correct retained answers per original problem. Savings compare aggregate generated tokens or wall time with paired full generation. Brackets give 95\% paired problem-bootstrap intervals for time savings. Target calibration coverage is 75\% and patience is one. Shading is for readability, not significance.}
\label{tab:online}
\end{table*}

\subsection{Vocabulary projection is a material cost}
For hidden width $d$, vocabulary size $V$, and $L$ translators, the dominant additional scalar-extraction terms at a check are approximately $2Ld^2$ for affine translation and $2LdV$ for vocabulary projection, counting multiplication and addition separately. Entropy and KL also require elementwise operations. Omitting the vocabulary term substantially understates cost even when the learned translator is small relative to the language model.

At a context of 512 tokens and checks every 16 tokens, the dense-FLOP estimate adds about 22.5\% for Qwen 3B and 13.6\% for Qwen 7B. Across the eight models it ranges from 4.2\% to 26.1\%. These estimates exclude normalization, rotary embedding, memory movement, and scheduling, and approximate elementwise arithmetic. They explain a potential cost mechanism but do not predict hardware latency exactly. Measured wall time remains the operational evidence.

\section{Checkpoints and Layer Indexing}
\label{app:models}
\begin{table}[h]
\centering\tablefont\setlength{\tabcolsep}{2.5pt}
\begin{tabular}{lrrrr}
\toprule
\textbf{Model} & \textbf{Problems} & \textbf{Correct (\%)} & \textbf{Mid index} & \textbf{Layers} \\
\midrule
\rowcolor{resultshade}
Qwen 1.5B & 500 & 38.6 & 17 & 28 \\
Qwen 3B & 500 & 64.0 & 22 & 36 \\
\rowcolor{resultshade}
Qwen 7B & 500 & 71.6 & 17 & 28 \\
Qwen 14B & 500 & 77.0 & 29 & 48 \\
\rowcolor{resultshade}
Mistral 7B & 500 & 15.0 & 19 & 32 \\
OLMo 7B & 500 & 33.4 & 19 & 32 \\
\rowcolor{resultshade}
Llama 8B & 500 & 43.8 & 19 & 32 \\
Yi 9B & 500 & 46.6 & 29 & 48 \\
\bottomrule
\end{tabular}

\caption{\textbf{Primary cached MATH-500 coverage.} Correct percentages use the current symbolic grader. Mid index counts completed transformer blocks, with embedding state indexed zero. The final state includes final normalization. These cached accuracies are distinct from the 100-problem online accuracies in Table~\ref{tab:online}.}
\end{table}
Qwen identifiers are Qwen/Qwen2.5-1.5B-Instruct, Qwen/Qwen2.5-3B-Instruct, Qwen/Qwen2.5-7B-Instruct, and Qwen/Qwen2.5-14B-Instruct. The remaining identifiers are mistralai/Mistral-7B-Instruct-v0.3, allenai/OLMo-2-1124-7B-Instruct, meta-llama/Llama-3.1-8B-Instruct, and 01-ai/Yi-1.5-9B-Chat. Source hashes in the extraction manifests identify the actual lens and response artifacts. These identifiers describe the tested checkpoints rather than implying that every family release behaves similarly.

\section{Sensitivity Definitions}
\label{app:sensitivity}
\begin{table*}[t]
\centering\tablefont\setlength{\tabcolsep}{4pt}
\begin{tabular}{lrr}
\toprule
\textbf{Model} & \textbf{Residual minus position} & \textbf{Five permutation differences} \\
\midrule
\rowcolor{resultshade}
Qwen 1.5B & $-0.26$ [-0.64, +0.14] & [+0.08, +0.19] \\
Qwen 3B & $-0.00$ [-0.18, +0.16] & [-0.12, +0.09] \\
\rowcolor{resultshade}
Qwen 7B & $-0.29$ [-0.54, -0.03] & [-0.34, -0.20] \\
Qwen 14B & $-0.26$ [-0.50, -0.04] & [-0.41, -0.01] \\
\rowcolor{resultshade}
Mistral 7B & $-1.81$ [-2.60, -1.01] & [-0.47, +0.23] \\
OLMo 7B & $-0.86$ [-1.19, -0.52] & [-0.13, +0.09] \\
\rowcolor{resultshade}
Llama 8B & $-0.12$ [-0.42, +0.21] & [+0.16, +0.20] \\
Yi 9B & $-0.24$ [-0.43, -0.04] & [-0.02, +0.09] \\
\bottomrule
\end{tabular}

\caption{\textbf{Residual and permutation sensitivity on MATH-500.} Residual minus position compares cross-fitted scalar and position-only corrections after selecting among four output penalties on tuning data, with the previously selected architecture held fixed. Values are AUROC points with individual paired 95\% intervals. The final column is the minimum and maximum real-hybrid-minus-permuted point estimate across five permutation draws, not a confidence interval.}
\end{table*}
The random-feature control has the same added dimensionality as the scalar block. Token-level permutation respects partition boundaries and causal position. Whole-problem permutation exchanges complete sequences only among problems with identical measured-position schedules. No-position controls remove the entire four-component position vector from the final and hybrid inputs. The capacity-matched intermediate control uses the final-state-selected architecture rather than independently choosing its width. These diagnostics answer different conditional questions and are not pooled into a single null test.

The main incremental family contains 32 comparisons. Raw-intermediate-only versus final comparisons are outside that family and are described with individual intervals. The five additional permutation draws are reported as sensitivity results rather than added retrospectively to a favorable testing family. Practical bounds use the same conditional bootstrap scope as the primary contrasts.

\begin{table*}[t]
\centering\tablefont\setlength{\tabcolsep}{2.2pt}
\begin{tabular}{lrrrr}
\toprule
\textbf{Model} & \textbf{Random added} & \textbf{Final no pos.} & \textbf{Hybrid no pos.} & \textbf{Matched mid} \\
\midrule
\rowcolor{resultshade}
Qwen 1.5B & $-0.16$ [-0.44, +0.11] & $-0.58$ [-0.91, -0.24] & $-0.41$ [-0.81, -0.02] & $+3.06$ [+1.70, +4.37] \\
Qwen 3B & $-0.05$ [-0.49, +0.37] & $-0.13$ [-0.54, +0.29] & $-0.73$ [-1.25, -0.21] & $+4.83$ [+3.57, +6.07] \\
\rowcolor{resultshade}
Qwen 7B & $+0.07$ [-0.10, +0.24] & $-0.56$ [-0.84, -0.27] & $-0.47$ [-0.86, -0.10] & $+4.95$ [+3.58, +6.28] \\
Qwen 14B & $-0.21$ [-0.64, +0.24] & $-0.27$ [-1.14, +0.61] & $+0.29$ [-0.17, +0.73] & $+3.69$ [+2.10, +5.13] \\
\rowcolor{resultshade}
Mistral 7B & $+1.15$ [+0.65, +1.65] & $+0.45$ [-0.21, +1.10] & $-0.27$ [-1.04, +0.54] & $+5.73$ [+2.96, +8.28] \\
OLMo 7B & $-0.18$ [-0.81, +0.45] & $-0.15$ [-0.57, +0.29] & $+0.16$ [-0.42, +0.76] & $+2.01$ [+0.61, +3.38] \\
\rowcolor{resultshade}
Llama 8B & $+0.51$ [+0.17, +0.86] & $+0.28$ [-0.24, +0.77] & $+0.54$ [+0.13, +0.97] & $+0.73$ [-0.41, +1.89] \\
Yi 9B & $+0.51$ [+0.25, +0.75] & $+0.25$ [-0.02, +0.49] & $+0.34$ [+0.08, +0.60] & $+1.10$ [-0.28, +2.52] \\
\bottomrule
\end{tabular}

\caption{\textbf{Additional sensitivity comparisons against the final-state probe with position.} Entries are AUROC point changes with individual paired 95\% problem-bootstrap intervals on MATH-500. Random added augments final states with dimensionality-matched random features. Final no pos. and Hybrid no pos. remove position inputs, but both displayed differences use the primary baseline with position, not each other. Matched mid uses the baseline-selected architecture. These are sensitivity results outside the primary 32-test family.}
\end{table*}

\begin{table*}[!t]
\centering\tablefont\setlength{\tabcolsep}{3.5pt}
\begin{tabular}{lrrr}
\toprule
\textbf{Model} & \textbf{Scalar minus final} & \textbf{Hybrid minus final} & \textbf{Hybrid minus permuted} \\
\midrule
\rowcolor{resultshade}
Qwen 3B & $-7.18$ [-9.10, -5.28] & $-0.08$ [-0.60, +0.43] & $+0.08$ [-0.32, +0.50] \\
Qwen 7B & $-6.47$ [-8.56, -4.44] & $+0.01$ [-0.24, +0.25] & $-0.05$ [-0.25, +0.15] \\
\rowcolor{resultshade}
Mistral 7B & $-3.49$ [-6.92, +0.05] & $+0.96$ [+0.28, +1.65] & $+0.11$ [-0.28, +0.51] \\
OLMo 7B & $-2.87$ [-5.24, -0.51] & $-0.17$ [-0.84, +0.52] & $-0.01$ [-0.26, +0.24] \\
\rowcolor{resultshade}
Llama 8B & $+0.07$ [-2.37, +2.38] & $+0.63$ [+0.27, +0.98] & $+0.16$ [+0.05, +0.26] \\
Yi 9B & $-8.61$ [-10.86, -6.45] & $+0.47$ [+0.20, +0.72] & $-0.01$ [-0.13, +0.11] \\
\bottomrule
\end{tabular}

\caption{\textbf{Math-trained lens sensitivity on the six available models.} Differences are problem-balanced token ROC-AUC points with individual paired 95\% intervals under the revised grading and evaluation protocol. Scalar and hybrid definitions match the main text. The permuted hybrid preserves causal position and final-state inputs. This table describes incremental prediction with these lenses, not a paired test of superiority over web-trained lenses.}
\end{table*}

The six math-lens comparisons retain small positive scalar and hybrid exceptions, especially for Llama. They do not justify a claim of universal underperformance. The separately retained reasoning-checkpoint extension gives a raw-intermediate gain of 3.50 AUROC points with interval 2.40 to 4.57, while the hybrid change is 0.23 points with interval minus 0.06 to 0.53. That extension is outside the eight-model family and is not independent confirmation on a new benchmark.

\section{Historical Reporting Corrections}
\label{app:corrections}
The earlier draft called ROC-AUC area under selective accuracy. Historical hybrid and residual code also normalized position by completed trace length, which is unavailable during generation. The revised fixed-budget comparison is a changed protocol and cannot be relabeled as the original experiment. Its feature scaling, tuning, cross-fitting, and grouped calibration likewise belong to the revised evaluation.

Historical regrading changes 412 incorrect labels to correct and one correct label to incorrect, giving 413 transitions and a net increase of 411. The previously reported 286 compared stored labels with a different recomputed checker. Neither number describes v6-versus-v4 label changes, which are zero on the nine current caches. The author packet separately contains the 125 current-grader disagreements with stored labels and 180 seeded agreements. These distinct comparisons should not share one heading or denominator.

Seven of seven signs yield exact two-sided probability 0.015625, and six of seven yield 0.125. Holm correction over the four historical domain sign tests gives 0.0625 for the three seven-of-seven tests. The earlier corrected-significance claim is therefore withdrawn. The historical math-lens table has two improvements, three deteriorations, and one tie at displayed precision. Earlier baseline values of 0.699, 0.708, and 0.763 belong to different grading, generation, and split regimes and are not interchangeable.

The historical Qwen 1.5B extension script generates MATH-500 only, explaining the seven-model coverage of its non-math table. Similar positive rates across benchmarks do not establish a causal domain effect. These historical results are not used to support a new claim that all trajectory gains disappear.

\section{Cost and Reproducibility Details}
\label{app:cost}
The completed campaign contains 15 missing-control extensions, eight sensitivity suites, 11 policy recalibrations, and eight online shards. Existing core fits were reused only after checking labels, problem identities, feature signatures, lengths, and evaluator hashes. The rejected v5 grading trial is explicitly excluded because inspection exposed false negatives on valid prose. The v6 regression suite and full protocol tests total 55 passing tests.

Execution used eight A100 80 GB GPUs and an existing Python 3.12 environment with PyTorch 2.6.0+cu124, Transformers 5.14.1, Math-Verify 0.9.0, NumPy 2.4.4, and scikit-learn 1.9.0. Models loaded from the existing cache. The online timing path uses PyTorch rather than vLLM. Aggregate wall time includes prefill and overhead but excludes model loading, probe fitting, and warm-up. Paired uncertainty holds the GPU assignment fixed. FLOP arithmetic includes attention projections, grouped-query dimensions where applicable, attention score/value products, gated feed-forward layers, final unembedding, affine lens translation, and every lens vocabulary projection. Elementwise cost is approximate.

\end{document}